\documentclass[sigconf,nonacm]{acmart}

\usepackage{amsmath}
\usepackage{booktabs}
\usepackage{tabularx}
\usepackage{array}
\usepackage{graphicx}
\usepackage{microtype}
\usepackage{enumitem}

\setcopyright{none}
\DeclareMathOperator*{\argmax}{arg\,max}
\DeclareMathOperator{\softmax}{softmax}
\DeclareMathOperator{\LN}{LN}

\newcolumntype{Y}{>{\centering\arraybackslash}X}

\newcommand{\egfn}{EEG--fNIRS}
\newcommand{\best}[1]{\textbf{#1}}

\newcommand{\FigurePlaceholder}[3]{%
  \fbox{%
    \begin{minipage}[c][#1][c]{%
      \dimexpr\linewidth-2\fboxsep-2\fboxrule\relax
    }
      \centering
      \small
      \textbf{#2}\par\medskip
      \begin{minipage}{0.92\linewidth}
        \centering #3
      \end{minipage}
    \end{minipage}%
  }%
}

\begin{document}

\title{Motor Imagery Classification Using Hybrid Cross-Subject Class Mixup and Selective Attention}

\author{Jiwon Kim}
\authornote{Both authors contributed equally to this research.}
\affiliation{%
  \institution{Yonsei University}
  \country{Republic of Korea}
}

\author{Chae-Won Lee}
\authornotemark[1]
\affiliation{%
  \institution{Yonsei University}
  \country{Republic of Korea}
}

\author{Jong-Seok Lee}
\affiliation{%
  \institution{Yonsei University}
  \country{Republic of Korea}
}

\begin{abstract}
Motor imagery (MI) enables a brain--computer interface (BCI) to decode movement intentions without overt movement. Hybrid electroencephalography (EEG) and functional near-infrared spectroscopy (fNIRS) classification combines complementary signals, but inter-subject variability and the design of information exchange between modalities affect generalization to new subjects. We propose a framework that combines Cross-Subject Class Mixup (CSCM) with Selective Attention. CSCM aims to construct new training data to make the model robust against inter-subject variability. It independently selects EEG and fNIRS donor trials from other training subjects within the anchor class and mixes each modality while preserving the original label. Selective Attention provides an effective way to simultaneously model the temporal signal characteristics and fuse the two modalities. It retains causal temporal self-attention for EEG and uses a classification token (CLS) to integrate both modalities, without fNIRS attention updates or direct cross-modal temporal-token attention. Leave-one-subject-out (LOSO) evaluation across 28 subjects on a public EEG--fNIRS dataset yields a mean balanced accuracy (BA) of 76.85\% and a macro-averaged F1 score (Macro-F1) of 76.58\%. The proposed model achieves the highest mean performance among five reimplemented baselines. We also examine the contributions of CSCM and Selective Attention through modality-specific mixing and donor-policy comparisons, temporal-policy comparisons, interventions on direct cross-modal paths, and component ablations.
\end{abstract}

\maketitle

\begin{figure}[t]
  \centering
  \IfFileExists{figures/fig1_cscm.pdf}{%
    \includegraphics[width=\linewidth]{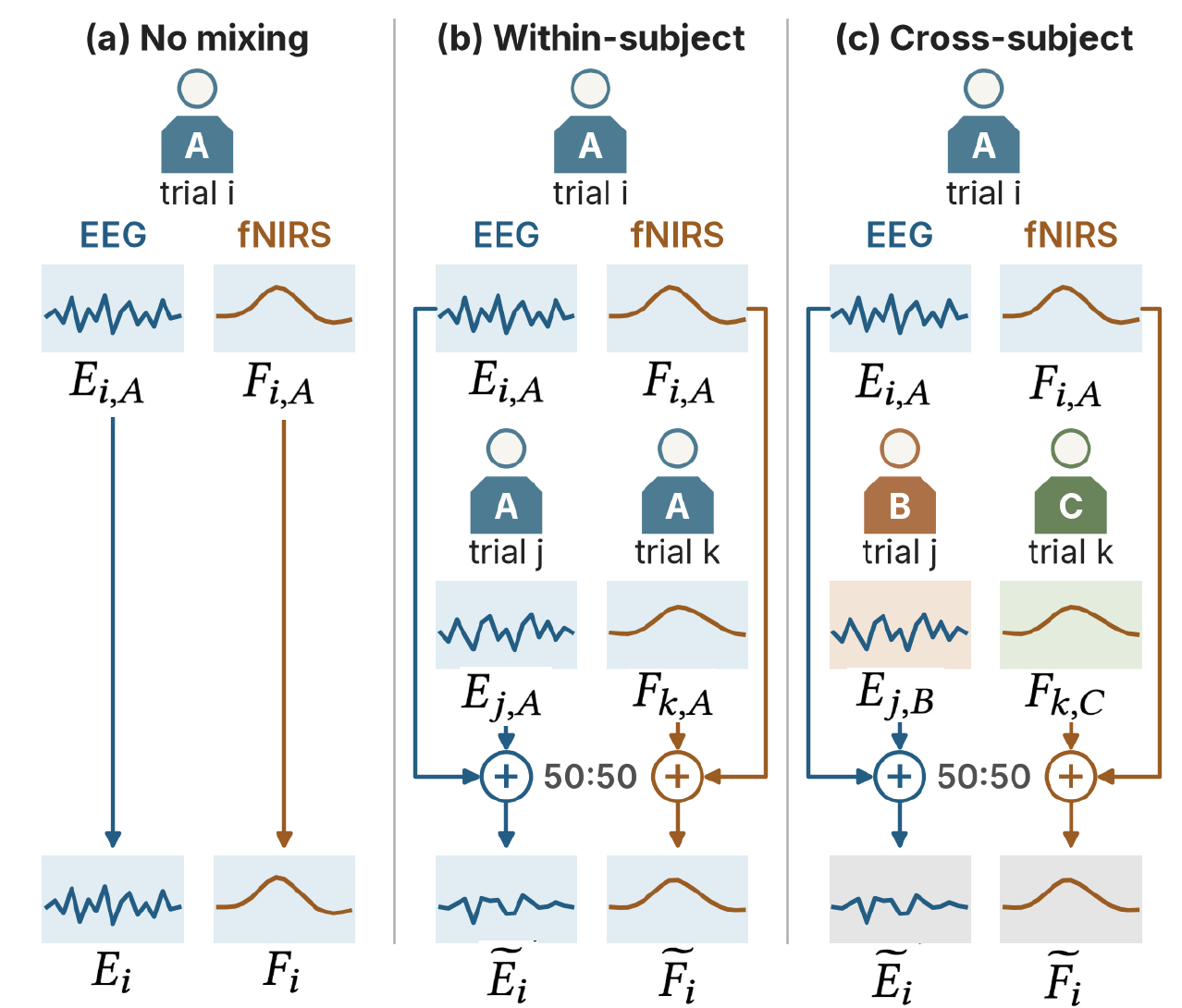}%
  }{%
    \FigurePlaceholder{1.22in}{CSCM: artwork pending}{%
      (a) No mixing \quad (b) Within-subject mixing\par
      (c) Same-class cross-subject mixing\par\smallskip
      Subject identities, class labels, independent EEG/fNIRS donors, and the unchanged target label.}%
  }
\caption{
EEG--fNIRS training-input construction:
(a) no mixing, (b) within-subject mixing, and (c) CSCM.
Mixing uses same-class donors at a 50:50 ratio and preserves
the anchor label. CSCM selects EEG and fNIRS donors independently.
Letters (A, B, C) denote subjects, and signal subscripts denote
trial (i, j, k) and subject identities, respectively.
}
\label{fig:cscm}

\Description{
Three-panel comparison of no mixing, within-subject mixing,
and cross-subject class mixup.
EEG and fNIRS signals are shown separately.
Within-subject donors belong to the anchor subject,
whereas cross-subject donors come from other training subjects.
Mixed signals use equal anchor and donor weights.
The anchor class label is preserved.
}
\end{figure}
\section{Introduction}
\label{sec:introduction}

A brain--computer interface (BCI) measures and analyzes neural signals to control a computer or external device, providing a communication pathway that does not depend on overt muscular movement~\cite{wolpaw2002}. Motor imagery (MI) is a BCI paradigm in which a user imagines a movement without physically performing it, and the corresponding brain activity is decoded~\cite{pfurtscheller2001mi}. Typical tasks include imagining the repeated clenching and opening of the left or right hand; finer-grained tasks concern wrist flexion and extension or individual finger movements. MI-based systems support communication and assistive control for people with movement or communication impairments, as well as robotic control and motor rehabilitation~\cite{wolpaw2002,ramos2013}.

Electroencephalography (EEG) and functional near-infrared spectroscopy (fNIRS) provide two noninvasive approaches to measuring brain activity during MI. EEG records electrical activity at millisecond-scale temporal resolution but is susceptible to noise and nonstationary signal statistics. fNIRS uses optical measurements, is less susceptible to electrical interference, and records a slower hemodynamic response that follows neural activity by several seconds~\cite{hong2018hybrid,efmlnet}. Hybrid \egfn\ classification uses this complementarity to improve performance relative to individual modalities~\cite{shin2017}. Its architectures range from concatenated handcrafted features to modality-specific branches, dedicated fusion modules, and attention-based deep models~\cite{hong2018hybrid,fganet,stanet,efmlnet}. As these architectures become more elaborate, it is also important to determine which components and information-exchange paths contribute to classification performance.

Meanwhile, EEG and fNIRS exhibit substantial inter-subject variability in signal characteristics and statistical distributions~\cite{ea,wang2021fnirs}. Models that perform well in subject-dependent settings can show lower generalization performance in subject-independent settings without subject-specific calibration~\cite{saha2020}. Large differences across subjects and the resulting difficulty in generalization are among the major obstacles to the real-world deployment of BCI systems. Improving BCI performance therefore requires not only high classification accuracy but also better generalization at both the data and model levels, so that the system operates robustly despite inter-subject differences.

We propose a hybrid \egfn\ MI classification framework that combines Cross-Subject Class Mixup (CSCM) and Selective Attention to improve cross-subject performance under leave-one-subject-out (LOSO) evaluation. At the data level, CSCM mixes real training signals from different subjects within the same class. EEG and fNIRS donors are selected independently, so the two modalities need not use the same donor trial or subject. Because each donor shares the anchor label, the mixed input retains that label. This construction exposes the model to combinations of signals from different subjects while preserving class membership, which can ultimately improve robustness under inter-subject variability. Figure~\ref{fig:cscm} contrasts the resulting input construction with no mixing and within-subject mixing.

At the model level, we investigate an configuration combining the two modalities. Specifically, our Selective Attention retains causal EEG temporal attention and allows a classification token (CLS) to read the information from both modalities. EEG tokens can read CLS, while fNIRS tokens retain residual and feed-forward processing and provide keys and values to CLS. Direct \egfn\ temporal-token connections and fNIRS attention updates are disabled. We examine these modality-specific policies and the effects of direct cross-modal paths through controlled comparisons.

Our framework achieves a balanced accuracy (BA) of 76.85\% and a macro-averaged F1 score (Macro-F1) of 76.58\% under matched 28-subject, three-seed LOSO evaluation. These are the highest mean scores among the five reimplemented baselines, exceeding EFMLNet, the strongest baseline in BA, by 3.81 percentage points. We additionally provide extensive experimental results supporting our design choices.

Our contributions are summarized as follows:
\begin{itemize}[leftmargin=8pt, topsep=0.8pt, itemsep=1pt, parsep=0pt]
  \item \textbf{Hybrid classification framework.} We propose an \egfn\ framework that combines CSCM and Selective Attention for cross-subject MI classification. It achieves 76.85\% BA and 76.58\% Macro-F1, exceeding the performance of the five baselines.
  \item \textbf{Cross-Subject Class Mixup (CSCM).} We construct training inputs by independently selecting same-class EEG and fNIRS donors from other training subjects. Modality-specific mixing and no-, within-, and cross-subject donor-policy comparisons assess this design.
  \item \textbf{Selective Attention.} We combine causal EEG temporal attention with CLS-based integration of both modalities. Temporal-policy comparisons and interventions on direct cross-modal paths examine the effectiveness of this routing.
\end{itemize}
\section{Related Work}
\label{sec:related}

\subsection{Brain--Computer Interfaces and Motor Imagery}

BCI acquisition is invasive or noninvasive. Invasive approaches place electrodes on or within neural tissue and require surgery and long-term safety management; noninvasive approaches record  brain activity signals from the scalp or head surface~\cite{wolpaw2002}. EEG and fNIRS belong to the latter group: EEG provides high temporal resolution, while fNIRS provides more localized hemodynamic measurements with less electrical interference and a slower response~\cite{hong2018hybrid,efmlnet}.

BCI paradigms use either external-stimulus responses, including steady-state visual evoked potentials and P300 responses, or self-generated tasks such as MI and mental arithmetic~\cite{wolpaw2002,shin2017}. MI is an endogenous paradigm that engages movement-related activity without overt movement, rather than relying on stimulus-evoked responses~\cite{pfurtscheller2001mi}. Tasks involve hands, feet, or individual joints; this work classifies left- versus right-hand MI. Voluntary sensorimotor activity supports the assistive, robotic, neuroprosthetic, and rehabilitation applications~\cite{wolpaw2002,ramos2013}. Variability across subjects and within subjects over time makes reliable decoding and cross-subject generalization central research problems~\cite{saha2020,ea}.

\subsection{\egfn\ Multimodal Architectures}

Traditional EEG-based MI decoding includes handcrafted features such as event-related desynchronization and synchronization, band power, and common spatial patterns, followed by classifiers such as linear discriminant analysis or support vector machines~\cite{shin2017}. Deep models instead learn features from signals or time--frequency representations using convolutional neural networks, recurrent neural networks, or Transformers~\cite{fganet,stanet,efmlnet}. Because EEG captures rapid changes, temporal dependencies and relationships between measurement channels are important aspects of its representation.

fNIRS classification uses changes in oxygenated hemoglobin (HbO) and deoxygenated hemoglobin (HbR). Feature-based approaches summarize these signals using means, slopes, and peaks, while deep approaches use convolutional, recurrent, or attention-based processing~\cite{hong2018hybrid,wang2021fnirs}. The slower hemodynamic response motivates representations of longer temporal intervals and the spatial distribution of activity across measurement locations.

Hybrid models combine these complementary signals at the feature, decision, or intermediate-representation level~\cite{hong2018hybrid,efmlnet}. Feature-level fusion concatenates features, decision-level fusion combines classifier outputs, and deep fusion integrates representations from modality-specific branches. Attention-based architectures model relationships both within each modality and between modalities~\cite{fganet,stanet,efmlnet}. Their design determines which temporal relationships and cross-modal information paths are available to the classifier.

As fusion architectures become more elaborate, it remains important to determine whether their interaction paths contribute to performance and cross-subject generalization, which is investigated in this paper.

\subsection{Signal Augmentation}

BCI recordings are costly to collect, while limited labeled trials and inter-subject signal variability motivate training-data augmentation~\cite{shin2017,rommel2022,ea,wang2021fnirs}. Approaches include noise injection, temporal cropping or perturbation, frequency-domain transformations, and synthetic signal generation~\cite{rommel2022,wickramaratne2021}. Mixup constructs convex combinations of training samples and labels, exposing the classifier to interpolated inputs as a regularization strategy for generalization~\cite{mixup}.

Wang \textit{et al.}~\cite{wang2021fnirs} apply cross-subject mixing to fNIRS-only \textit{n}-back workload classification. Their procedure selects corresponding temporal windows from different subjects with a common task order, and uses a coefficient derived from a beta distribution. Mixing is used in the pretraining phase.

CSCM extends this same-class cross-subject mixing principle to hybrid MI classification using independently selected EEG and fNIRS donors.

\begin{figure*}[t]
  \centering
  \IfFileExists{figures/fig2_framework.pdf}{%
    \includegraphics[width=\textwidth]{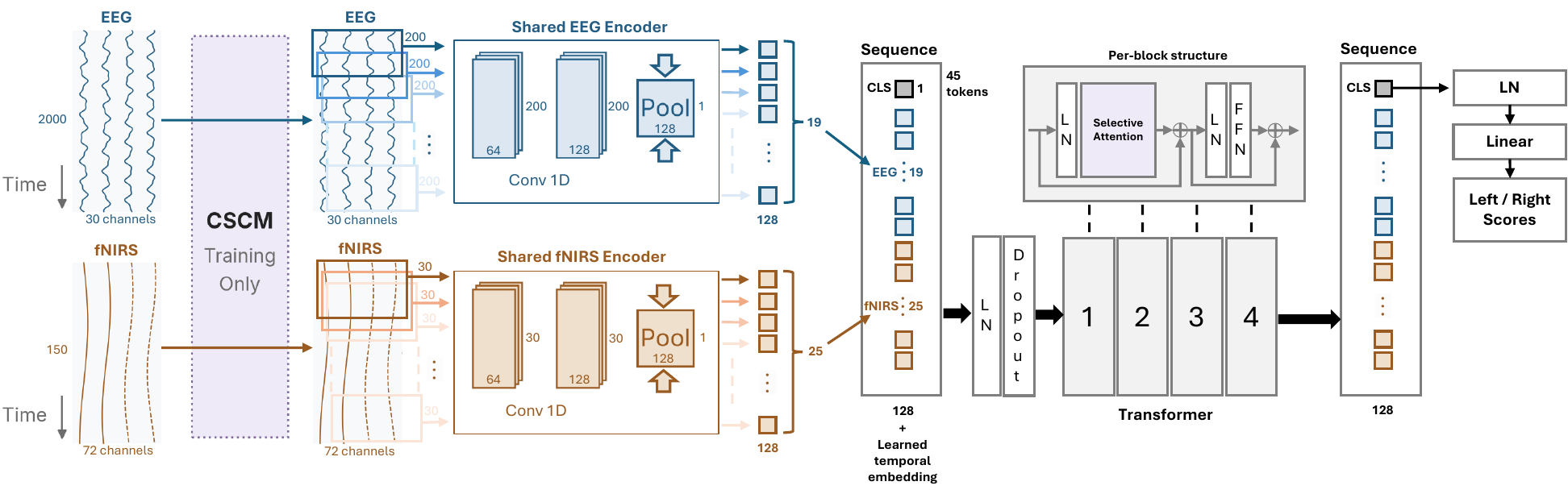}%
  }{}

  \vspace{-0em}

  \caption{
    Overview of the proposed framework. CSCM is applied only during
    training, while inference uses the original paired signals.
    The modality-specific encoders produce 19 EEG and 25 fNIRS tokens,
    forming a 45-token sequence with CLS. Learned temporal embeddings
    are added before layer normalization, dropout, and four Transformer
    blocks. The final CLS representation is used for left/right MI
    classification.
  }
  \label{fig:framework}
  \Description{
    Architecture diagram with two input streams, training-only CSCM,
    independent temporal encoders, a 45-token sequence, four Transformer
    blocks, and a final CLS classifier.
  }
\end{figure*}
\section{Problem Formulation}
\label{sec:problem}
\label{sec:task} 

We consider binary left- versus right-hand MI classification using paired EEG and fNIRS signals. The two modalities have different measurement channels and temporal resolutions, so each trial is represented by respective matrices:
\begin{equation}
  E_i\in\mathbb{R}^{C_E\times T_E},\qquad
  F_i\in\mathbb{R}^{C_F\times T_F}.
  \label{eq:inputs}
\end{equation}
Here, $i$ is the trial index and $\mathbb{R}$ is the set of real numbers. $C_E$ and $C_F$ are the EEG and fNIRS channel counts, while $T_E$ and $T_F$ are their numbers of temporal samples, respectively. The entries $E_i(c,t)$ and $F_i(c,t)$ denote the values at channel index $c$ and sample index $t$. Because the sampling rates differ, one sample interval represents a different duration in the two modalities. The preprocessed input sizes are $30\times2000$ for EEG and $72\times150$ for fNIRS.

The dataset is denoted as
\begin{equation}
  \mathcal D=\{(E_i,F_i,y_i,s_i)\}_{i=1}^{N},
  \label{eq:dataset}
\end{equation}
where $N$ is the number of trials, $E_i$ and $F_i$ are signals from the same trial, $y_i\in\{0,1\}$ is its left/right MI label, and $s_i$ is its subject identifier.

The classifier maps the paired signals to two class scores and a predicted class:
\begin{equation}
  z_i=g_\theta(E_i,F_i)\in\mathbb{R}^{2},\qquad
  \hat y_i=\argmax_{r\in\{0,1\}} z_{i,r}.
  \label{eq:prediction}
\end{equation}
The function $g_\theta$ is the model with learned weights $\theta$, $z_i$ is its score vector, $r$ indexes the two classes, and $z_{i,r}$ is the score for class $r$. The prediction $\hat y_i$ is the class with the largest score.

For cross-subject evaluation, training and evaluation data are separated by subject. A subject's trials do not appear in both partitions. The model learns from paired signals and labels of the training subjects and predicts the MI class of a new subject. It receives no target-subject label supervision or classifier fine-tuning. Section~\ref{sec:setup} specifies the evaluation split and preprocessing conditions.

The objective is to learn from multiple subjects' paired signals and accurately classify left/right MI in a test subject. In particular, we pose two research questions: how EEG and fNIRS information can be integrated through a structure that accounts for their different temporal and informational characteristics (RQ1), and how training data can be constructed to improve classification on new subjects despite inter-subject signal variability (RQ2). Selective Attention addresses RQ1, and CSCM addresses RQ2. Section~\ref{sec:method} defines both methods, and Section~\ref{sec:experiments} evaluates their effects.

\section{Proposed Method}
\label{sec:method}

\subsection{Framework Overview}
\label{sec:overview}
\label{sec:objective} 

The proposed framework combines CSCM and Selective Attention for cross-subject MI classification. Figure~\ref{fig:framework} shows two modality-specific temporal encoders, a Transformer for information integration, and a classifier operating on the final CLS representation. During training, the encoders receive signals mixed by CSCM; during inference, they receive the original paired signals. Their outputs are concatenated with CLS and processed by the Transformer before classification.

Each temporal encoder partitions its input into overlapping windows and converts each window into a fixed-dimensional vector using one-dimensional convolution (Conv1D). The EEG encoder processes a 10-second signal at 200\,Hz with a 200-sample window and 100-sample stride, producing 19 windows. The fNIRS encoder processes a 15-second signal at 10\,Hz with a 30-sample window and five-sample stride, producing 25 windows.

Both encoders contain two Conv1D layers. For EEG, we use channel dimensions $30\rightarrow64\rightarrow128$ and kernel sizes 5 and 3. For fNIRS, we use $72\rightarrow64\rightarrow128$ and kernel sizes 15 and 7. Each convolution is followed by a Gaussian error linear unit (GELU). Adaptive average pooling over time within each window, layer normalization (LN), and dropout produce one 128-dimensional token.

The encoder outputs are
\begin{equation}
  \begin{aligned}
  H_i^E&=\operatorname{Encoder}_E(E_i)\in\mathbb{R}^{19\times128},\\
  H_i^F&=\operatorname{Encoder}_F(F_i)\in\mathbb{R}^{25\times128}.
  \end{aligned}
  \label{eq:encoders}
\end{equation}
Here, $\operatorname{Encoder}_E$ and $\operatorname{Encoder}_F$ are the respective encoders, and $H_i^E$ and $H_i^F$ contain their temporal tokens. Each token represents all input channels within one window. During training, the arguments in Equation~\eqref{eq:encoders} are the mixed signals defined in Section~\ref{sec:cscm}.

A learned CLS embedding is added to form
\begin{equation}
  X_i=[\mathrm{CLS};H_i^E;H_i^F]\in\mathbb{R}^{45\times128},
  \label{eq:sequence}
\end{equation}
where the semicolon denotes concatenation along the token axis. Therefore, the 45 tokens comprise one CLS, 19 EEG tokens, and 25 fNIRS tokens. Learned temporal-index embeddings are added to the token sequence, followed by LN and dropout.

The embedded sequence passes through four pre-LN Transformer blocks~\cite{xiong2020preln}. Each block contains multi-head attention and a feed-forward network, both with residual connections. Attention uses eight heads with dimension 16 per head. The feed-forward dimensions are $128\rightarrow256\rightarrow128$. The blocks use Selective Attention for modality-specific temporal processing and CLS-based integration, with the routes defined in Section~\ref{sec:selective}.

Let $h_i^{\mathrm{CLS}}\in\mathbb{R}^{128}$ be the CLS representation from the last block. The class scores are obtained by
\begin{equation}
  z_i=W_{\mathrm{cls}}\LN(h_i^{\mathrm{CLS}})+b_{\mathrm{cls}},
  \label{eq:classifier}
\end{equation}
where $W_{\mathrm{cls}}\in\mathbb{R}^{2\times128}$ and $b_{\mathrm{cls}}\in\mathbb{R}^{2}$ are the linear layer's weight matrix and bias. The final classification follows Equation~\eqref{eq:prediction}.

For training, cross-entropy at the final output, $L_{\mathrm{CE}}$, is combined with temporal regularization of adjacent EEG token representations, $L_{\mathrm{temp}}$:
\begin{equation}
  \mathcal L=\mathcal L_{\mathrm{CE}}
  +\lambda_{\mathrm{temp}}\mathcal L_{\mathrm{temp}}.
  \label{eq:loss}
\end{equation}
Here, $\lambda_{\mathrm{temp}}$ weights the temporal penalty given by
\begin{equation}
\begin{split}
  \mathcal L_{\mathrm{temp}}
  ={}&\frac{1}{B(T'_E-1)d}
  \sum_{b=1}^{B}\sum_{t=1}^{T'_E-1}\sum_{k=1}^{d}\left(H^{E,(4)}_{b,t+1,k}-H^{E,(4)}_{b,t,k}\right)^2.
\end{split}
  \label{eq:temporal-loss}
\end{equation}
$H^{E,(4)}\in\mathbb{R}^{B\times T'_E\times d}$ denotes the EEG token representations after the fourth Transformer block. $B$ is the batch size, $T'_E=19$ is the number of EEG tokens, and $d=128$ is the token dimension. The indices $b$, $t$, and $k$ identify the batch sample, the first token of an adjacent pair, and the feature dimension, respectively. The loss averages squared feature differences over the 18 adjacent pairs and all batch samples. It penalizes abrupt changes between adjacent EEG representations. We set $\lambda_{\mathrm{temp}}=0.01$.

\subsection{Cross-Subject Class Mixup}
\label{sec:cscm}

Even within one MI class, EEG and fNIRS characteristics significantly differ across subjects. CSCM incorporates combinations of signals from different training subjects while preserving the anchor label. Unlike unrestricted mixup, which interpolates inputs and target labels~\cite{mixup}, CSCM selects only donors with the anchor class.

We distinguish three training-input policies by donor selection. \emph{No mixing} uses the original signals. \emph{Within-subject class mixing} uses a different trial with the same subject and class as the anchor. \emph{Cross-subject class mixing}, which defines CSCM, uses a trial with the same class but a different subject. Figure~\ref{fig:cscm} illustrates these policies.

For the CSCM policy, the eligible donor set for anchor trial $i$ is
\begin{equation}
  \mathcal J_i=\{j\in\mathcal I_{\mathrm{train}}
    \mid y_j=y_i,\ s_j\ne s_i\}.
  \label{eq:donors}
\end{equation}
The set $\mathcal I_{\mathrm{train}}$ contains trial indices from the current training fold, and $j$ denotes a candidate donor index. The constraints require same class labels and different subject identities. Donor selection therefore uses only the current training subjects.

The EEG donor index $j$ and fNIRS donor index $k$ are sampled independently from $\mathcal J_i$. The two donors are not required to come from the same trial or subject, although both are possible. Each must satisfy the class and subject constraints relative to the anchor. For example, the anchor EEG can be mixed with EEG from subject 5 and the anchor fNIRS with fNIRS from subject 7, provided both donors satisfy these constraints.

Given these donors, the mixed inputs are
\begin{equation}
  \begin{aligned}
    \widetilde E_i&=w_EE_i+(1-w_E)E_j,\\
    \widetilde F_i&=w_FF_i+(1-w_F)F_k.
  \end{aligned}
  \label{eq:mixing}
\end{equation}
The anchor signals are $E_i$ and $F_i$, the donor signals are $E_j$ and $F_k$, and the resulting inputs are $\widetilde E_i$ and $\widetilde F_i$. The coefficients $w_E,w_F\in[0,1]$ control the relative contributions of the anchor and doner.

Both donors have the anchor class, so the mixed pair retains $y_i$. By presenting combinations of same-class signals from different training subjects, this construction is intended to reduce reliance on subject- or trial-specific patterns. Section~\ref{sec:cscm-results} evaluates its effect through modality-specific mixing and donor-policy comparisons.

\subsection{Selective Attention}
\label{sec:selective}

EEG and fNIRS measure different physiological processes and exhibit different temporal characteristics. Selective Attention therefore assigns different roles to the two modalities. Causal temporal attention is applied to EEG tokens, while fNIRS tokens bypass attention updates but remain available to CLS. The classification token (CLS) integrates information from both modalities, and EEG tokens can access CLS. This design examines whether fNIRS representations can support multimodal classification through CLS without additional temporal attention updates or direct EEG--fNIRS temporal-token attention.

We implement masked multi-head attention based on scaled dot-product attention~\cite{vaswani2017attention}. For each attention head, the attention scores and outputs are computed as
\begin{equation}
  \begin{aligned}
    S&=\frac{\LN(Q)\LN(K)^\top}{\sqrt{d_h}}+M,\\
    A&=\softmax(S),\qquad O=AV.
  \end{aligned}
  \label{eq:masked-attention}
\end{equation}
Here, $Q$, $K$, and $V$ are the query, key, and value matrices obtained from the input token representations. LN normalizes each query and key over its head dimension $d_h=16$. The mask $M$ determines which tokens can exchange information: permitted connections have value 0, while blocked connections have value $-\infty$. Softmax is applied along the key dimension, and the outputs of all attention heads are concatenated and projected.

Selective Attention controls both the permitted attention connections and the application of attention updates. Let $\mathbf c$ denote CLS, $\mathbf e_t$ an EEG token, and $\mathbf f_t$ a fNIRS token, where $t$ indicates the temporal position within each modality. For a query token $q$, let $\mathcal A(q)$ denote the set of tokens available as keys and values.

CLS attends to all EEG and fNIRS tokens, excluding itself:
\begin{equation}
  \mathcal A(\mathbf c)
   =\{\mathbf e_1,\ldots,\mathbf e_{19},
      \mathbf f_1,\ldots,\mathbf f_{25}\}.
  \label{eq:cls-routes}
\end{equation}
Each EEG token attends to CLS and its current and preceding EEG tokens, using a causal mask for direct within-EEG temporal attention:
\begin{equation}
  \mathcal A(\mathbf e_t)
   =\{\mathbf c,\mathbf e_1,\ldots,\mathbf e_t\},
  \quad t\in\{1,\ldots,19\}.
  \label{eq:eeg-routes}
\end{equation}
In contrast, fNIRS tokens do not receive attention updates. The projected attention update is set to zero before residual addition:
\begin{equation}
  U_{\mathrm{attn}}(\mathbf f_t)=\mathbf 0,
  \quad t\in\{1,\ldots,25\},
  \label{eq:fnirs-update}
\end{equation}
where $U_{\mathrm{attn}}$ denotes the projected multi-head attention update and $\mathbf 0\in\mathbb R^{128}$ is the zero vector. This bypass disables fNIRS attention updates while retaining residual and feed-forward processing. The resulting fNIRS representations remain available to CLS for multimodal integration.

Direct attention connections between EEG and fNIRS temporal tokens are disabled. Nevertheless, fNIRS information can reach EEG indirectly through CLS across successive Transformer blocks.

\begin{figure}[t]
  \centering
  \IfFileExists{figures/fig3_routing.pdf}{%
    \includegraphics[width=\linewidth]{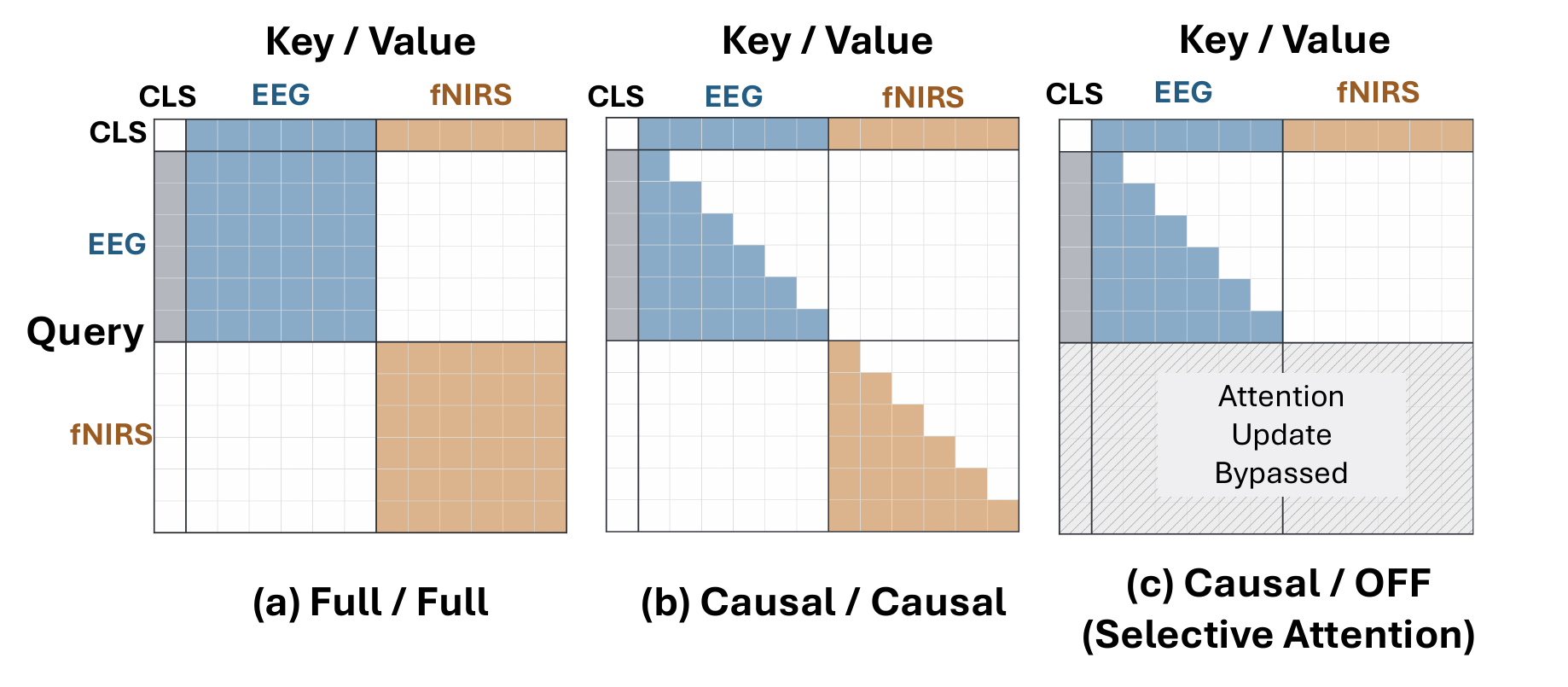}%
  }{%
    \FigurePlaceholder{1.55in}{Temporal routing: artwork pending}{%
      (a) EEG Full / fNIRS Full\qquad
      (b) EEG Causal / fNIRS Causal\par\smallskip
      (c) EEG Causal / fNIRS OFF}%
  }
\caption{
  Representative temporal-attention configurations with
  fixed CLS-based integration:
  (a) Full/Full, (b) Causal/Causal, and
  (c) Causal/OFF (Selective Attention).
  Rows represent queries and columns represent keys and values.
  The 19 EEG and 25 fNIRS tokens are shown schematically.
  CLS attends to both modalities, while EEG tokens can attend to CLS.
  Selective Attention bypasses fNIRS attention updates
  while retaining residual and feed-forward processing.
  Direct cross-modal temporal-token attention is disabled throughout.
}
  \label{fig:routing}
  \Description{
    Three attention-routing matrices comparing Full/Full,
    Causal/Causal, and Causal/OFF configurations.
    Rows represent queries and columns represent keys and values.
    Selective Attention bypasses fNIRS attention updates
    while retaining fNIRS representations for CLS-based integration.
  }
\end{figure}

Figure~\ref{fig:routing} illustrates three representative EEG/fNIRS temporal-attention configurations: Full/Full, Causal/Causal, and Causal/OFF (Selective Attention). All three retain CLS-based integration and disable direct cross-modal temporal-token attention.
\section{Experiments}
\label{sec:experiments}

We first compare the framework with existing hybrid classifiers. We then examine modality-specific mixing, donor policies, temporal attention, and interventions on direct cross-modal paths. Finally, we evaluate the separate and combined effects of the two components and sensitivity to the mixing weights.

\subsection{Experimental Setup}
\label{sec:setup}

\noindent\textbf{Dataset.}
We use the public hybrid dataset of Shin et al.~\cite{shin2017}, which provides simultaneous EEG and fNIRS recordings for MI. Our task is binary left- versus right-hand MI. Of the original 29 subjects, S01 is excluded because quality control identifies nonpositive raw fNIRS intensity values. The evaluation includes S02--S29, totaling 28 subjects, each with 30 trials per class and 60 trials overall. The complete dataset contains 1,680 trials.

\noindent\textbf{Data Preprocessing.}
For EEG, we use 30 channels, apply common average referencing, and filter each complete recording session with a third-order, 8--30\,Hz Butterworth band-pass filter in zero-phase mode. Subject-wise Euclidean alignment (EA) aligns covariance characteristics~\cite{ea}. The 0--10\,s interval of each trial at 200\,Hz forms a $30\times2000$ input.

For fNIRS, raw light intensity is converted to HbO and HbR concentration changes using the modified Beer--Lambert law (MBLL). A third-order, 0.01--0.09\,Hz zero-phase Butterworth band-pass filter is applied to each complete session. Baseline correction uses the interval from $-5$ to $-2$\,s. The model receives the 0--15\,s interval at 10\,Hz, with 36 measurement positions represented by 72 HbO/HbR channels. Selected motor-region positions undergo spatial normalization, and unused positions remain zero-filled. The resulting input has size $72\times150$.

\noindent\textbf{Evaluation.}
For cross-subject evaluation, each LOSO fold trains on 1,620 trials from 27 subjects and evaluates 60 trials from the remaining subject. All 28 subjects serve once as the held-out subject. The subject-specific EA covariance estimate uses each subject's unlabeled EEG trials, including all unlabeled trials of the held-out subject. No held-out labels are used for training or classifier fine-tuning.

Performance is evaluated using BA as the primary metric and Macro-F1 as the secondary metric. BA averages recall across the two classes:
\begin{equation}
  \mathrm{BA}=\frac12\left(
    \frac{\mathrm{TP}}{\mathrm{TP}+\mathrm{FN}}+
    \frac{\mathrm{TN}}{\mathrm{TN}+\mathrm{FP}}\right).
  \label{eq:ba}
\end{equation}
where $\mathrm{TP}$ and $\mathrm{TN}$ count correct positive and negative predictions, respectively. $\mathrm{FP}$ counts negative samples predicted as positive, and $\mathrm{FN}$ counts positive samples predicted as negative.

Macro-F1 averages the two class-specific F1 scores:
\begin{equation}
  \mathrm{Macro\mbox{-}F1}
    =\frac12\sum_{c=0}^{1}\frac{2P_cR_c}{P_c+R_c}.
  \label{eq:f1}
\end{equation}
Here, $c$ indexes the classes. Precision $P_c$ is the proportion of samples predicted as class $c$ whose true label is $c$, whereas recall $R_c$ is the proportion of samples with true label $c$ that are predicted as class $c$. Both metrics are reported as percentages. The three seed results are averaged within each held-out subject, followed by a mean over 28 subjects.

For statistical comparisons, we used two-sided exact paired sign-flip tests on subject-level differences, averaging the three seeds within each of the 28 held-out subjects. Holm correction was applied to five pairwise comparisons reported in the main-results and component-ablation analyses, separately for BA and Macro-F1, with a significance level of $\alpha=0.05$.

\noindent\textbf{Training Details.}
Training uses AdamW~\cite{loshchilov2019adamw} and the settings in Table~\ref{tab:training}, with each LOSO fold and seed trained independently. The scheduler uses a linear warm-up followed by a cosine schedule. Because its 150-epoch horizon exceeds the actual 20-epoch training duration, only the warm-up and the initial cosine segment are used. The first epoch uses one fifteenth of the base learning rate. Evaluation uses the final exponential moving average (EMA) checkpoint from this fixed training budget, with the loss in Equation~\eqref{eq:loss}.

For a fair comparison, CSCM is applied to the training inputs of both the proposed model and all five baselines. Each baseline retains its original architecture and training objectives. Ablations modify only the mixing or attention settings specified in their respective sections.

\subsection{Main Results}
\label{sec:main-results}

Table~\ref{tab:main} compares our framework with E-FNet~\cite{efnet}, CMD3~\cite{cmd3}, STA-Net~\cite{stanet}, FGANet~\cite{fganet}, and EFMLNet~\cite{efmlnet} under the same LOSO splits and three random seeds specified in Section~\ref{sec:setup}. Our framework achieves 76.85\% BA and 76.58\% Macro-F1, the highest mean scores among the compared models. It exceeds EFMLNet, the strongest baseline in BA, by 3.81 percentage points in BA and 4.05 percentage points in Macro-F1. Both differences are statistically significant after Holm correction ($p_{\mathrm{adj}}=0.00119$ for BA and $p_{\mathrm{adj}}=0.00102$ for Macro-F1).

\begin{table}[t]
  \caption{Training settings of the proposed model.}
    \vspace{-0em}
  \label{tab:training}
  \centering\small
  \begin{tabularx}{\columnwidth}{@{}l>{\raggedright\arraybackslash}X@{}}
    \toprule
    Setting & Value \\
    \midrule
    Optimizer & AdamW \\
    Base learning rate & $5\times10^{-4}$ \\
    Weight decay & $10^{-3}$ \\
    Batch size & 32 \\
    Training epochs & 20 \\
    Scheduler & LambdaLR; warm-up and cosine \\
    Warm-up & 15 epochs \\
    Scheduler horizon & 150 epochs \\
    Gradient clipping & Maximum norm 1.0 \\
    Dropout & 0.3 (encoder output, embedding, and Transformer) \\
    EMA start & Epoch 16 \\
    EMA decay & 0.995 \\
    Evaluation checkpoint & Final EMA at epoch 20 \\
    Temporal loss weight & 0.01 \\
    Random seeds & 1, 2, 3 \\
    \bottomrule
  \end{tabularx}
  \vspace{-0em}
\end{table}
\begin{table}[t]
  \caption{Classification performance under matched three-seed LOSO evaluation. Results are reported as mean $\pm$ SD across three seeds, with each seed averaged over 28 held-out subjects.}
  \label{tab:main}
  \centering\small
    \vspace{-0em}
  \begin{tabular}{@{}lcc@{}}
    \toprule
    Model & BA (\%) & Macro-F1 (\%) \\
    \midrule
    E-FNet & $61.59 \pm 5.72$ & $55.43 \pm 9.23$ \\
    CMD3 & $68.91 \pm 0.74$ & $67.16 \pm 1.44$ \\
    STA-Net & $64.62 \pm 2.34$ & $63.56 \pm 2.20$ \\
    FGANet & $69.52 \pm 1.05$ & $69.04 \pm 1.09$ \\
    EFMLNet & $73.04 \pm 1.35$ & $72.53 \pm 1.53$ \\
    \best{Ours} & $\boldsymbol{76.85 \pm 0.27}$ & $\boldsymbol{76.58 \pm 0.27}$ \\
    \bottomrule
  \end{tabular}
\end{table}

\subsection{Effect of CSCM}
\label{sec:cscm-results}

CSCM involves both mixing training trials and selecting donors from other subjects. To separate these two choices, we compare no mixing, within-subject class mixing, and cross-subject class mixing for each modality. We evaluate all combinations of three mixing policies (no mixing, within-subject mixing, and cross-subject mixing) and three temporal-attention policies (OFF, Causal, and Full) independently for EEG and fNIRS, yielding $3^4=81$ configurations. Table~\ref{tab:donor-policy}(a) reports the nine EEG/fNIRS donor-policy combinations, each averaged over all nine combinations of EEG and fNIRS temporal-attention policies. Each mixed modality uses an anchor weight of 0.5, and evaluation uses the original paired EEG and fNIRS inputs in every condition.

\begin{table}[t]
  \caption{
    Donor-policy performance (BA / Macro-F1, \%) across
    the full 81-configuration factorial study.
    (a) Each EEG/fNIRS donor-policy combination is averaged
    over nine temporal-attention configurations.
    (b) Each modality's donor policy is averaged over
    the other modality's donor policy and both temporal-attention
    policies (27 configurations).
    Direct cross-modal attention is OFF throughout.
  }
  \label{tab:donor-policy}
  \centering\small
    \vspace{-0em}
  \textbf{(a) Two-factor donor-policy averages}\par\smallskip
  \setlength{\tabcolsep}{3pt}
  \begin{tabularx}{\columnwidth}{@{}lYYY@{}}
    \toprule
    EEG / fNIRS & No mixing & Within-subject & Cross-subject \\
    \midrule
    No mixing
      & 72.82 / 72.43
      & 75.02 / 74.67
      & 75.30 / 74.95 \\
    Within-subject
      & 73.50 / 73.14
      & 74.89 / 74.49
      & 75.43 / 75.10 \\
    Cross-subject
      & 74.45 / 74.10
      & 75.72 / 75.40
      & \best{75.86 / 75.58} \\
    \bottomrule
  \end{tabularx}

  \par\medskip
  \textbf{(b) Single-factor donor-policy averages}\par\smallskip
  \begin{tabularx}{\columnwidth}{@{}lYY@{}}
    \toprule
    Donor policy & EEG mixing & fNIRS mixing \\
    \midrule
    No mixing
      & 74.38 / 74.02
      & 73.59 / 73.22 \\
    Within-subject
      & 74.61 / 74.24
      & 75.21 / 74.86 \\
    Cross-subject
      & \best{75.35 / 75.03}
      & \best{75.53 / 75.21} \\
    \bottomrule
  \end{tabularx}
\end{table}

Across the attention configurations, applying cross-subject mixing to both modalities increases mean BA from 72.82\% to 75.86\% and Macro-F1 from 72.43\% to 75.58\%. Mixing either modality alone also improves mean performance over no mixing, with a larger increase from fNIRS-only mixing. Mixing both yields the highest mean BA and Macro-F1. Comparing within-subject and cross-subject mixing further reveals whether selecting donors from other subjects provides an additional benefit beyond mixing within the same subject. For each fixed policy in one modality, cross-subject donors give higher mean BA than within-subject donors. The size of this increase depends on the other modality's policy.

Table~\ref{tab:donor-policy}(b) separately summarizes the donor-policy effects for EEG and fNIRS. For each fixed donor policy in one modality, the results are averaged over the other modality's three donor policies and both modalities' temporal-attention policies ($3^3=27$ configurations). These values also correspond to the row and column averages of Table~\ref{tab:donor-policy}(a). Cross-subject mixing achieves the highest marginal BA and Macro-F1 for both modalities, followed by within-subject mixing and no mixing.

\subsection{Effect of Temporal and Cross-Modal Attention}
\label{sec:attention-results}
\label{sec:intervention} 
\begin{table}[t]
    \caption{
      Direct cross-modal temporal-token attention interventions
      with EEG temporal attention fixed to Causal.
      Scores are BA (\%).
      $\Delta$ is the change from each model family's original score
      (Orig.) in percentage points.
      Each row averages the three fNIRS temporal-attention policies
      (OFF, Causal, and Full) over three seeds and 28 subjects.
    }
  \label{tab:temporal}
  \centering\small
    \vspace{-0em}
  \setlength{\tabcolsep}{3pt}
  \begin{tabularx}{\columnwidth}{@{}lYYY@{}}
    \toprule
    EEG / fNIRS & OFF & Causal & Full \\
    \midrule
    OFF
      & 75.40 / 75.14
      & 75.58 / 75.29
      & 75.56 / 75.24 \\
    Causal
      & \best{76.85 / 76.58}
      & 76.39 / 76.11
      & 76.71 / 76.47 \\
    Full
      & 75.42 / 75.07
      & 75.75 / 75.47
      & 75.14 / 74.82 \\
    \bottomrule
  \end{tabularx}
\end{table}
\begin{table}[t]
  \caption{Direct cross-modal path interventions. Scores are BA (\%); $\Delta$ is the change from the original score (Orig.) in percentage points. Each row averages its direction-specific model family over three fNIRS temporal policies, three seeds, and 28 subjects.}
  \label{tab:intervention}
  \centering\small
  \vspace{-0em}
  \begin{tabular*}{\columnwidth}{@{\extracolsep{\fill}}lrrrrr@{}}
    \toprule
    & & \multicolumn{2}{c}{Masking} & \multicolumn{2}{c}{Value removal} \\
    \cmidrule(lr){3-4}\cmidrule(l){5-6}
    Removed path & Orig. & BA & $\Delta$ & BA & $\Delta$ \\
    \midrule
    fNIRS $\rightarrow$ EEG & 76.47 & 76.07 & $-0.40$ & 76.26 & $-0.22$ \\
    EEG $\rightarrow$ fNIRS & 76.58 & 76.65 & $+0.07$ & 76.59 & $+0.01$ \\
    Bidirectional & 76.59 & 76.31 & $-0.28$ & 76.24 & $-0.34$ \\
    \bottomrule
  \end{tabular*}
\end{table}

We first examine how the temporal-attention policies assigned to EEG and fNIRS affect classification performance. With CSCM applied to both modalities, we evaluate all nine combinations of OFF, Causal, and Full temporal attention for EEG and fNIRS. All nine configurations retain CLS readout, with the CLS token attending to both EEG and fNIRS tokens. Direct cross-modal temporal-token attention remains OFF. Causal restricts attention between temporal tokens to the current and preceding positions within each modality, whereas Full allows attention to all temporal positions. Table~\ref{tab:temporal} reports BA and Macro-F1 for these nine configurations.

Across all three fNIRS temporal-attention policies, causal EEG attention achieves the highest mean BA and Macro-F1. Based on this consistent pattern, we retain causal EEG attention in the subsequent experiment.

We next test how classification performance changes when direct EEG--fNIRS attention paths are disabled during evaluation of trained models. We use separately trained models with EEG-to-fNIRS, fNIRS-to-EEG, or bidirectional attention paths and keep their learned parameters fixed, without retraining. Here, fNIRS-to-EEG denotes fNIRS queries attending to EEG keys and values, whereas EEG-to-fNIRS denotes EEG queries attending to fNIRS keys and values. EEG attention remains causal, and each direction-specific model family includes fNIRS temporal policies OFF, Causal, and Full.

We remove the selected attention paths in two ways. \emph{Attention-path masking} blocks the selected query--key pairs and recomputes the softmax in Equation \eqref{eq:masked-attention} over the remaining permitted keys. \emph{Value-contribution removal} keeps the original forward pass's attention weights but removes the selected weighted-value terms before output projection. Values and hidden states are recomputed under this intervention, while the retained attention weights are not renormalized. Both interventions act on the same paths in all four blocks.

Table~\ref{tab:intervention} reports the change from each model family's own original score. Results average the three fNIRS temporal-attention policies (OFF, Causal, and Full) and three seeds within each subject, followed by the mean over 28 subjects. Masking fNIRS-to-EEG or bidirectional attention paths reduces mean BA, whereas masking EEG-to-fNIRS gives a small increase. Value-contribution removal also produces small mean changes. The largest decrease under either intervention is 0.40 percentage points, indicating a small effect on mean BA when these attention paths are disabled during evaluation.

\subsection{Additional Analysis}
\label{sec:component}
\label{sec:weights}
\begin{table}[t]
    \caption{
      Component ablation of CSCM and Selective Attention.
      Full denotes unrestricted within-modality temporal attention
      in both EEG and fNIRS.
    }
  \label{tab:component}
  \centering\small
    \vspace{-0em}
  \begin{tabular}{@{}llrr@{}}
    \toprule
    CSCM & Attention & BA (\%) & Macro-F1 (\%) \\
    \midrule
    OFF & Full temporal & 72.56 & 72.09 \\
    ON & Full temporal & 75.14 & 74.82 \\
    OFF & Selective Attention & 73.79 & 73.50 \\
    \best{ON} & \best{Selective Attention} & \best{76.85} & \best{76.58} \\
    \bottomrule
  \end{tabular}
\end{table}
\begin{table}[t]
  \caption{Mixing-weight sensitivity. Each entry reports BA (\%) / Macro-F1 (\%). Rows and columns specify the EEG and fNIRS anchor weights.}
  \label{tab:weights}
  \centering\small
    \vspace{-0em}
  \setlength{\tabcolsep}{3pt}
  \begin{tabularx}{\columnwidth}{@{}lYYY@{}}
    \toprule
    $w_E$ / $w_F$ & 0.25 & 0.50 & 0.75 \\
    \midrule
    0.25 & 75.85 / 75.51 & 75.97 / 75.65 & 75.97 / 75.59 \\
    0.50 & 76.37 / 76.13 & \best{76.85 / 76.58} & 76.45 / 76.20 \\
    0.75 & 76.57 / 76.35 & 76.23 / 75.96 & 75.14 / 74.80 \\
    \bottomrule
  \end{tabularx}
\end{table}

To determine whether each component improves performance with and without the other, Table~\ref{tab:component} reports a $2\times2$ ablation comparing CSCM ON/OFF under full temporal attention or Selective Attention. Full temporal attention enables all within-modality temporal connections in both modalities. All four conditions retain CLS-based integration, keep direct cross-modal paths OFF, and share the remaining architecture and evaluation protocol. CSCM increases BA by 2.58 percentage points with full temporal attention and by 3.06 percentage points with Selective Attention. Both improvements are statistically significant for BA and Macro-F1 after Holm correction (all $p_{\mathrm{adj}}<0.003$). Conversely, Selective Attention increases BA over full temporal attention by 1.23 percentage points without CSCM and by 1.71 percentage points with CSCM. Its Macro-F1 improvements are statistically significant both without CSCM ($p_{\mathrm{adj}}=0.03645$) and with CSCM ($p_{\mathrm{adj}}=0.02363$). The BA improvement is significant with CSCM ($p_{\mathrm{adj}}=0.02909$), but not without CSCM ($p_{\mathrm{adj}}=0.05081$). Each component improves mean BA and Macro-F1 both with and without the other, and their combination achieves the highest mean scores.

Finally, we examine whether the performance improvement from CSCM depends on the anchor-to-donor mixing ratio. We vary $w_E$ and $w_F$ independently, while keeping cross-subject donor selection and Selective Attention fixed. Table~\ref{tab:weights} presents nine combinations of EEG and fNIRS anchor weights using values of 0.25, 0.50, and 0.75. Rows give EEG anchor weights and columns give fNIRS anchor weights. All nine settings achieve higher mean BA and Macro-F1 than the CSCM-OFF condition under the same attention policy (73.79\% BA and 73.50\% Macro-F1 in Table~\ref{tab:component}). Among these nine settings, the adopted equal-weight configuration, $w_E=w_F=0.5$, achieves the highest mean BA and Macro-F1.

\section{Discussion}
\label{sec:discussion}

The CSCM results suggest that the subject supplying a donor signal matters in addition to whether mixing is applied. With the other modality's policy fixed, cross-subject donors yield higher mean BA than within-subject donors, although the increase varies with that policy. CSCM combines signals from multiple subjects within the same class without changing the target label. Presenting these combinations during training can act as a regularizer against reliance on individual trial patterns.

The attention-policy comparison shows that causal EEG attention achieves the highest mean BA and Macro-F1 for each fNIRS temporal-attention policy, with Causal/OFF yielding the highest mean scores among the nine configurations. Selective Attention retains the fNIRS encoder, residual and feed-forward representations, and CLS readout while using causal temporal attention for EEG. This preserves fNIRS as an input to the classifier even when its attention updates are disabled. In separate evaluations with model parameters fixed, masking direct cross-modal attention paths or removing their weighted-value contributions produces only small changes in mean BA in the evaluated model families. Together, these results support modality-specific temporal processing within the evaluated CLS-based integration framework.

\balance

\section{Conclusion}
\label{sec:conclusion}

We proposed a hybrid MI classification framework that combines CSCM and Selective Attention. CSCM independently mixes each modality with same-class donor signals from other training subjects while preserving the anchor label. Selective Attention uses causal EEG temporal attention while disabling fNIRS temporal-attention updates, with both modalities integrated through CLS.

The present evaluation covers one public dataset and binary left- versus right-hand MI. Evaluation on other subject populations, datasets, and tasks in the future will be valuable to establish how widely the two designs apply.

\bibliographystyle{ACM-Reference-Format}
\bibliography{ref}
\end{document}